\documentclass[letterpaper,twocolumn,fleqn]{article} 
\usepackage{ist}
\usepackage{cite}
\usepackage{hyperref}
\usepackage{color}
\title{Automatic Facial Skin Feature Detection for Everyone}

\author{Qian Zheng$^1$, Ankur Purwar$^2$, Heng Zhao$^1$, Guang Liang Lim$^1$, Ling Li$^1$, Debasish Behera$^2$, Qian Wang$^1$, Min Tan$^1$, Rizhao Cai$^1$, Jennifer Werner$^2$, Dennis Sng$^1$, Maurice van Steensel$^1$, Weisi Lin$^1$, Alex C Kot$^1$;\\
$^1$ Nanyang Technological University, Singapore;\\
$^2$ Procter \& Gamble;
}
\date{} 

\begin{document} 
\maketitle 
\thispagestyle{empty} 


\begin{abstract}
Automatic assessment and understanding of facial skin condition have several applications, including the early detection of underlying health problems, lifestyle and dietary treatment, skin-care product recommendation, etc. 
Selfies in the wild serve as an excellent data resource to democratize skin quality assessment, but suffer from several data collection challenges.
The key to guaranteeing an accurate assessment is accurate detection of different skin features.
We present an automatic facial skin feature detection method that works across a variety of skin tones and age groups for selfies in the wild. 
To be specific, we annotate the locations of acne, pigmentation, and wrinkle for selfie images with different skin tone colors, severity levels, and lighting conditions. 
The annotation is conducted in a two-phase scheme with the help of a dermatologist to train volunteers for annotation. 
We employ Unet++ as the network architecture for feature detection.  
This work shows that the two-phase annotation scheme can robustly detect the accurate locations of acne, pigmentation, and wrinkle for selfie images with different ethnicities, skin tone colors, severity levels, age groups, and lighting conditions.
\end{abstract}

\section{Introduction}
\label{sec:intro}
 
Understanding facial skin quality is beneficial to beauty and health.
For mild to moderate cases, a trip to a dermatologist may not be worth the time and cost. 
Yet, when presented with the huge variety of off-the-shelf skin products, a consumer may be unsure of which is best for her needs.
An automatic facial skin quality assessment system not only saves time and money, but also helps protect privacy. 
Successfully detecting different facial skin features is the key to guaranteeing a comprehensive and accurate assessment of skin quality.
However, existing solutions either fail to identify problematic regions and only output overall grading ({\it e.g.}, learning-based methods~\cite{Xiaoping2019,Karunanayake2020}, commercial Apps~\cite{app1} or products~\cite{link1}) or are sensitive to unexpected factors such as lighting conditions and skin tone colors ({\it e.g.}, traditional methods~\cite{Fazly2016,Nasim2016,Thanapha2015,Maroni2017,Tsung2019}). 
Besides, all existing solutions ({\it e.g.},~\cite{Kyungseo2021}) deal with images captured in controlled environments, which limits its practicability in various scenarios for casual users. 
In contrast, this paper relaxes the constraint of controlled environments and targets on selfie images captured in the wild. 
Besides, this paper highlights the problematic regions for a deep understanding of facial skin quality for skin features of acne, pigmentation, and wrinkles, with the input of a single selfie image.

We regard the skin feature detection problem as a dense prediction problem~\cite{Simon2021}, {\it i.e.}, it produces a binary mask that pixel-wisely indicates the locations of acne, pigmentation, or wrinkle. 
There are many similar dense prediction problems in computer vision, such as semantic segmentation~\cite{Hyeonwoo2015}, image enhancement~\cite{Jong-Sen1980}, intrinsic image decomposition~\cite{Qifeng2013}, etc.
Different from these problems that exploit priors of significant color differences or/and the analytic image formation models, the color difference between regions of normal skin and interested skin are minor and the existing analytic image formation models cannot be directly used for the problem of skin feature detection.
From the view of image signal processing, the challenges of this problem are brought by the unexpected lighting conditions of selfies images and various skin tone colors.

To address the above challenges, the basic idea of this paper is to leverage the data-driven approach and learn priors from training data.
That is, by observing the locations of acne, pigmentation, and wrinkle for large-scale selfie images in the wild (with different lighting conditions and skin tone colors), our model is expected to learn robust features for the detection task.
Therefore, the key is to obtain training data with reliable annotation.
We observe that it is hard to obtain accurate annotation of acne and pigmentation with one-phase manual annotation due to the impact of image quality and lighting.
We also observe that the distribution of annotated wrinkle location is long-tailed~\cite{Alpheus2011}, {\it e.g.}, there are few wrinkles in regions of the forehead or crow's feet while they frequently appear in regions of under-eye and cheeks. 
To improve the annotation accuracy for acne and pigmentation, and obtain wrinkle data with balanced distribution, we adopt a two-phase scheme to annotate training data with the help of a dermatologist.
To be specific, a coarse annotation of the locations of acne, pigmentation, and wrinkle is first conducted on 3,755 face images. All these images are elaborately prepared for the needed diversity, with different skin tone colors, levels of severity, and conditions of lighting. 
With these images, we train the preliminary skin feature detection models. 
In the second phase, we conduct a fine-grained annotation based on the model-generated results of acne and pigmentation and get 1,994 refined annotations. 
We also annotate the wrinkle of 995 elder faces as the additional training data.
These data are used to improve the models of acne, pigmentation, and wrinkle. 
To summarize, the contributions of this work are as follows:
\begin{itemize}
\item We propose the first deep learning based skin feature detection approach for a single selfie image in the wild. We show that the proposed method achieves a superior performance advantage as compared with conventional methods.
\item We propose a two-phase scheme annotation method for large-scale selfie images in the wild. We show this scheme can produce reliable annotations to train deep learning based skin feature detection models. 
\end{itemize}

\section{Related Work}

As there are only a few works that address three skin features, {\it i.e.}, acne, pigmentation, and wrinkle, simultaneously.
We briefly review the advances for these three skin features, respectively.

\noindent{\bf Acne.}
The substantial advances of single image-based solutions for acne lesion analysis have been made for different tasks
, including the grading of severity~\cite{B2011,Xiaoping2019,Sophie2019}, sub-type classification~\cite{Mohammad2018,Xiaolei2018}, smoothness detection~\cite{CL2018}, and acne detection.
Existing solutions of acne detection are traditional methods, which employ the techniques of image signal processing for the task of acne detection, such as discrete wavelet~\cite{Fazly2016}, color clustering~\cite{Nasim2016}, and color space conversion~\cite{Thanapha2015,Natchapol2016}.
These methods are sensitive to environment changes and face textures.

\noindent{\bf Pigmentation.}
As one of the dominant features for pigmentation analysis is color, researchers introduce appearance/biophysical models to measure the distributions of melanin and hemoglobin. 
The appearance models consider characteristics of wavelength-dependent
scattering and absorption~\cite{Hisashi1998}, principle chromatic component~\cite{Norimichi2003}, and multi-layered structure~\cite{Symon1999}.
The biophysical models consider hyper-spectral responses of human skin to realistically render skin images by taking into account light diffusion scattering between different layers~\cite{Craig2005}, the amount of oil, melanin, and hemoglobin in skin~\cite{Donner2006}, the distributions of melanin and hemoglobin for different facial expressions~\cite{Jorge2010}, concentrations of chromophores~\cite{Jose2015}, and hyper-spectral surface and subsurface scattering effects of skin appearance~\cite{TF2015}.
These methods fail to take the impact of lighting colors into account, which may produce inaccurate results.

\noindent{\bf Wrinkle.}
Existing solutions of wrinkle detection are traditional methods.
These methods can be categorized into texture-based~\cite{NG2014,Batool2012}, filter-based~\cite{ChCh2015,Cula2013,Batool2015}, and shape model-based~\cite{Tsalakanidou2010,Huang2020,Umirzakova2021} methods.
Texture-based methods might fail for temporary wrinkles with nonlinear and blurry shapes.
Filter-based approaches are sensitive to lighting changing because shadow can weaken the edge information.
Shape model-based methods need lots of computation to fit the face landmarks.
 
Existing skin feature detection methods are traditional methods.
They are sensitive to lighting changes, face textures, and skin tone colors.
In this paper, we propose the first deep learning based skin feature detection method that deals with the input of a single selfie image in the wild with different lighting conditions and skin tone colors, for skin features of acne, pigmentation, and wrinkle.

\section{Methodology}

To train and test our skin feature detection models, we collect selfie images from consumers.
All these data are captured at home by casual users. 
Therefore, the lighting conditions are uncontrolled.
We manually exclude images with poor quality ({\it e.g.}, images with missing patches, extreme poses, very dark environment, low resolution, severely blurry) and get 3,755 images.
All these images are from females and with ethnicities of Asian, African, Caucasian, Indian, and Hispanic.

\subsection{Manual Annotation}
\begin{figure}[!hb]
\centering
  \includegraphics[width=1\linewidth]{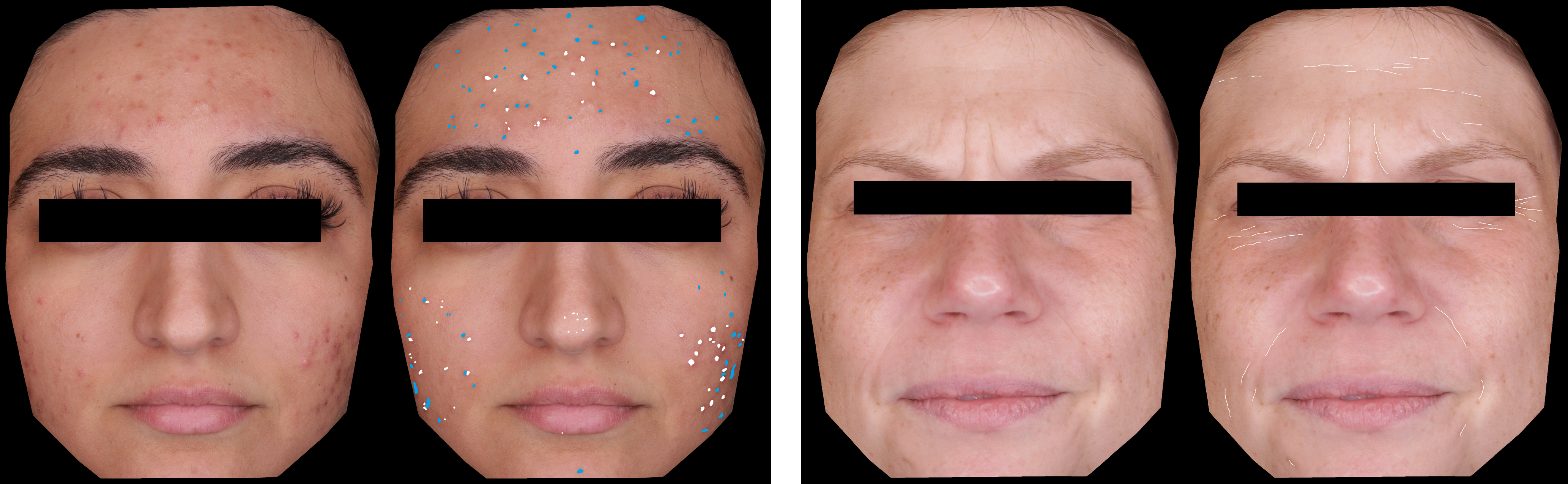}
  \caption{Left: the original image and its annotation result of acne (white dots) and pigmentation (blue dots), red dots represent uncertain annotation and we use them to train both acne and pigmentation detection models. Right: the original image and its annotation result of wrinkle. The eyes of face images are blocked for visualization due to privacy issues. }
  \label{fig:01}
\end{figure}

To mitigate the impact from the environment, we mask out the background based on Dlib~\cite{dlib}, an open-source library that can detect the facial profile. 
Our dermatologist then trains 4 volunteers to perform data annotation. 
The volunteers learn to discriminate acne, pigmentation, and wrinkle under different lighting conditions for faces with different skin colors. 
All volunteers are in the age group of 18-25 and are required to pass color-blind testing~\footnote{\url{https://enchroma.com/pages/color-blindness-test}} before the annotation.
The annotation result is represented as a mask image.
We annotate the masks of acne and pigmentation together due to that they are unlikely to appear at the same location.
Finally, 3,755 images are annotated to train our models.
Figure~\ref{fig:01} shows some examples of the annotation results.

\subsection{Two-Phase Scheme Annotation}

\begin{figure}[!ht]
\centering
  \includegraphics[width=0.7\linewidth]{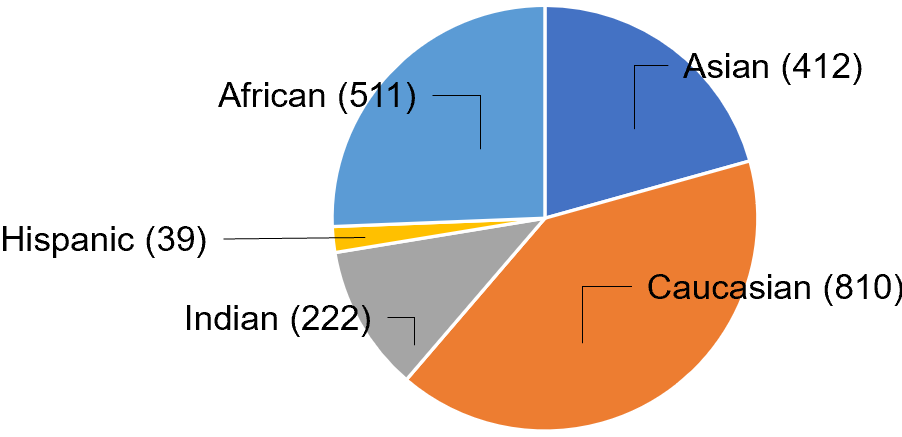}
  \caption{The image number of faces with each ethnicity in the training data.}
  \label{fig:0}
\end{figure}

We find that annotations of acne and pigmentation from different volunteers can hardly achieve agreement due to the impact of image quality and lighting. 
Therefore, we regard the first-phase annotations of acne and pigmentation as the coarse annotation and conduct the second-phase annotation for annotation refinement.
We then train two preliminary models for acne and pigmentation, respectively, using coarse annotated data as the training data.
The method is introduced in the next section.
We then conduct annotation refinement on these machine-generated data. 
The refinement is based on the machine-generated data instead of the those in the first phase for two reasons: 1) overall annotations are reliable so that the machine-generated results are accurate, {\it i.e.}, the accuracy of the machine-generated annotations are comparable or even better than the manually annotated ones, 2) machine-generated annotations are more consistent than manual ones.
1,994 annotations are refined in the second phase in total.
We then use these data to train our networks and get improved models of acne and pigmentation detection, respectively.
The number of images for each ethnicity is displayed in Figure~\ref{fig:0}.
The method used is introduced in the next section.

\begin{figure}[t]
\centering
  \includegraphics[width=0.8\linewidth]{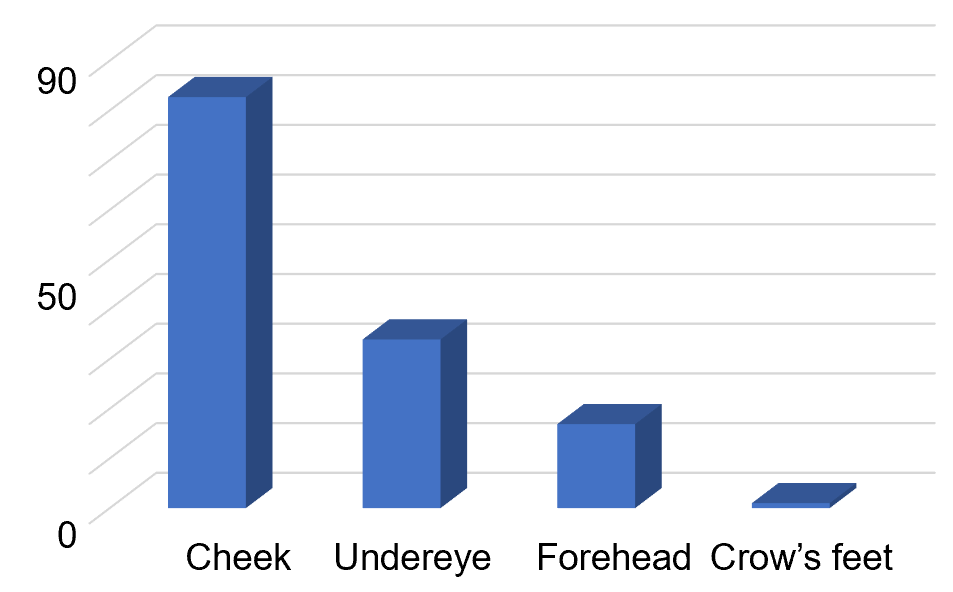}
  \caption{The percentage of valid annotation for wrinkle in terms of four facial regions ({\it i.e.,} cheek, undereye, forehead, crow's feet) in the first phase annotation.}
  \label{fig:3}
\end{figure}

\begin{figure}[t]
\centering
  \includegraphics[width=1\linewidth]{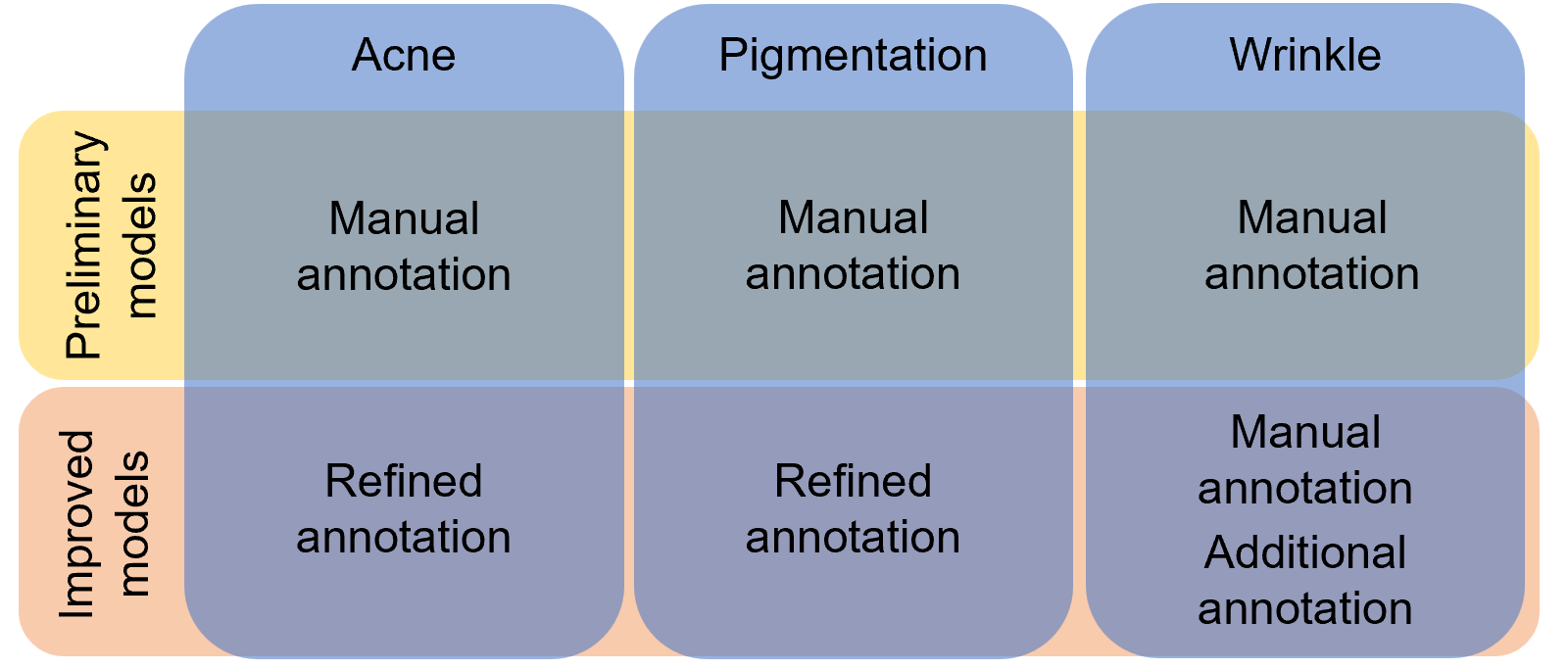}
  \caption{An overview of our two-phase scheme annotation.}
  \label{fig:1}
\end{figure}

We also train a wrinkle detection model based on the training data annotated in the first phase. 
However, we find the wrinkles on the crow's feet and forehead cannot be successfully detected due to the long-tailed distribution of our training data. 
Figure~\ref{fig:3} shows the distribution of the wrinkles in this training data.
Therefore, we additionally annotate 1,000 more selfies images from elder faces in the second phase.
Finally, we use data from both phases, 3,200 images in total, to train the wrinkle detection model.

The overview of our two-phase scheme annotation can be found in Figure~\ref{fig:1}.

\subsection{Data-Driven Method}

With the annotated data, the skin feature detection problem can be regarded as predicting the binary mask from the input of a selfie image.
Inspired by the great success of UNet++ for the problem of medical image segmentation~\cite{Zongwei2019}, we adopt the network architecture of UNet++ as the backbone to train our models.
As compared with other backbones, Unet++ not only alleviates the unknown network depth but also aggregates features of varying semantic scales. 
Deep supervision and attention~\cite{Zongwei2019} are also integrated into our method. 

We train three models for three skin features of acne, pigmentation, and wrinkle, respectively. 
To train the preliminary models, we use 3,500 out of 3,755 images for training.
For the remaining 255 images, we carefully select 90 of them, with different lighting conditions, severity, and skin tone colors as the testing set.
The loss function is to measure the similarity between the predicted annotation $\mathbf{\hat{M}}$ and the ground truth $\mathbf{M}$, which is implemented by an L1 loss,
\begin{equation}
\centering{
    \mathcal{L}=\|\mathbf{\hat{M}}-\mathbf{M}\|_1.
}
\end{equation}

The network architecture is based on that in~\cite{Zongwei2019}. All selfies images are cropped and resized to the size of $384\times384$.
The batch sizes are set to 16. 
All three models are trained using Adam solver with $\beta_1 = 0.9$ and
$\beta_2 = 0.999$.
We set the initial learning rates as 0.001. 


\section{Performance}

As there is seldom a study of using learning-based methods to solve the problem of skin feature detection, we compare performance with that from traditional methods.
To be specific, we compare the acne detection method from~\cite{Thanapha2015}, the pigmentation detection method from~\cite{lin2019exemplar}, and the wrinkle detection method from~\cite{Wrinkles_detection}. 
We use IoU as the metric for quantitative evaluation. 
Given the manually annotated/refined mask $\mathbf{M}$ and the predicted one $\mathbf{\hat{M}}$, the IoU is calculated by
\begin{equation}
    {IoU}(\mathbf{M},\mathbf{\hat{M}}) = \frac{sum({and}(\mathbf{M},\mathbf{\hat{M}}))}{sum({or}(\mathbf{M},\mathbf{\hat{M}}))},
\end{equation}
where `and' and `or' are the logical operators, `sum' is the summation operation. 

\begin{table}[!h]
\caption{The quantitative comparison in terms of IoU.}
\label{tab:1}
\begin{center}    
\begin{tabular}{c|c|c|c} 
\hline
Skin  & Preliminary  & Improved  & Traditional  \\
features&Models&Models&Methods\\
\hline
Acne& 0.1247 & 0.2300 & 0.1557\\
\hline
Pigmentation& 0.1624 & 0.3035 & 0.1348\\
\hline
Wrinkle& 0.0968 & 0.1512 & 0.1022\\
\hline
\end{tabular}
\end{center}
\end{table}

\begin{figure*}[t]
\centering
  \includegraphics[width=.8\linewidth]{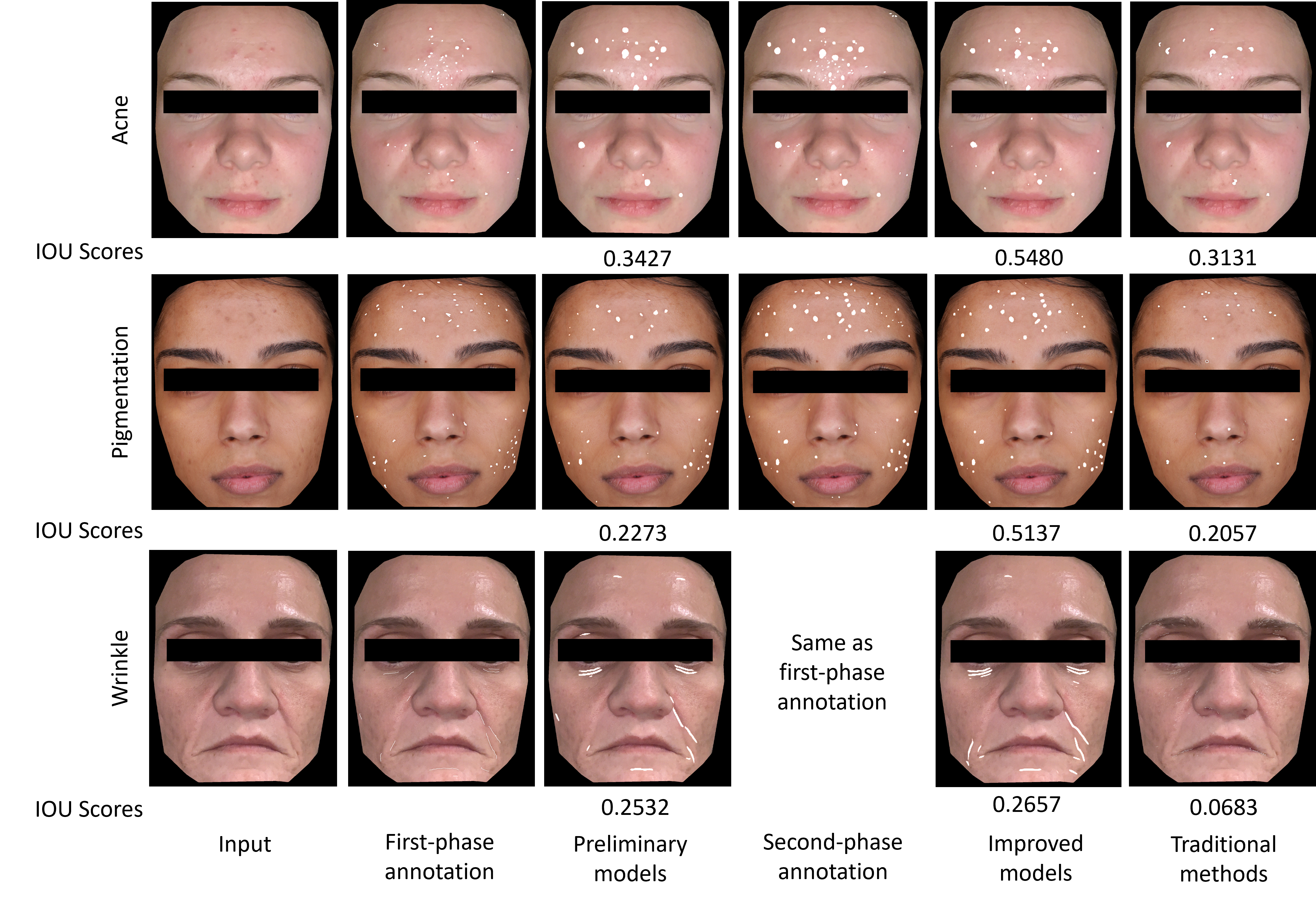}
  \caption{The performance comparison. From left to right: the input, annotations in first-phase, results from our preliminary models, annotations in the second phase, results from our improved models, and results from traditional methods. The numbers below images are IoU results. 
  The eyes of face images are blocked for visualization due to privacy issues. }
  \label{fig:4}
\end{figure*}

\begin{figure*}[t]
\centering
  \includegraphics[width=0.8\linewidth]{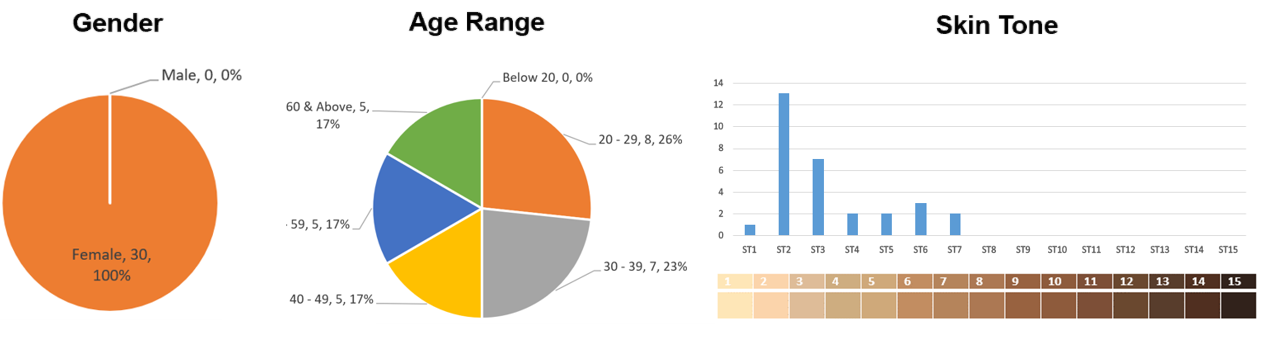}
  \caption{The respondents' information in terms of gender, age group, and skin tone color.}
  \label{fig:6}
\end{figure*}

As shown in Table~\ref{tab:1}, our method also produce the best IoU as compared with traditional methods.
Figure~\ref{fig:4} illustrates the visual comparison. 
As compared with traditional methods, results from our method are much more accurate as the performance of traditional methods are significantly impacted by unexpected patterns, lighting conditions, and skin tone colors. 
Note that the numbers are not high because these features are sparsely distributed.

We also compare detection results produced by our preliminary and improved models.
As can be found from Table~\ref{tab:1} and Figure~\ref{fig:4}, the models trained by data annotated in the second phase produce much more accurate predictions.

\section{Perception Study}
To further investigate the performance of our method, we conduct a perception study on people in the street. 
Following the advice from domain experts, we align the gender and ethnicity of the respondents and those of the testing images.
That is, both respondents and testing faces are Asian females.
We ask the respondents to provide their age and skin tone based on a scale between ST1 (lightest) and ST15(darkest).
The information of respondents can be found in Figure~\ref{fig:6}. 

We used 20 images for each skin condition in this study. 
The numbers of images for different severity of skin conditions are balanced, {\textit i.e.}, 7 for mild, 7 for moderate, and 6 for severe. 
All respondents are briefed on the basic knowledge of the types and severity of skin conditions.
The respondents are required to score the accuracy of our method's prediction results for each testing sample.
Two questions are asked for each image, {\it i.e.}, ``What do you think of the overall detection result?'', ``Are there any incorrectly detected acne?'', to evaluate the algorithm performance in terms of false-negative error and false-positive error, respectively.
There are three optional scores for each question, {\it i.e.}, bad, good, very good for the first question, and none, a few, many for the second question.
We also ask the respondents to grade the overall accuracy of our method for all images for each skin condition in a third question.

\begin{figure}[t]
\centering
  \includegraphics[width=1\linewidth]{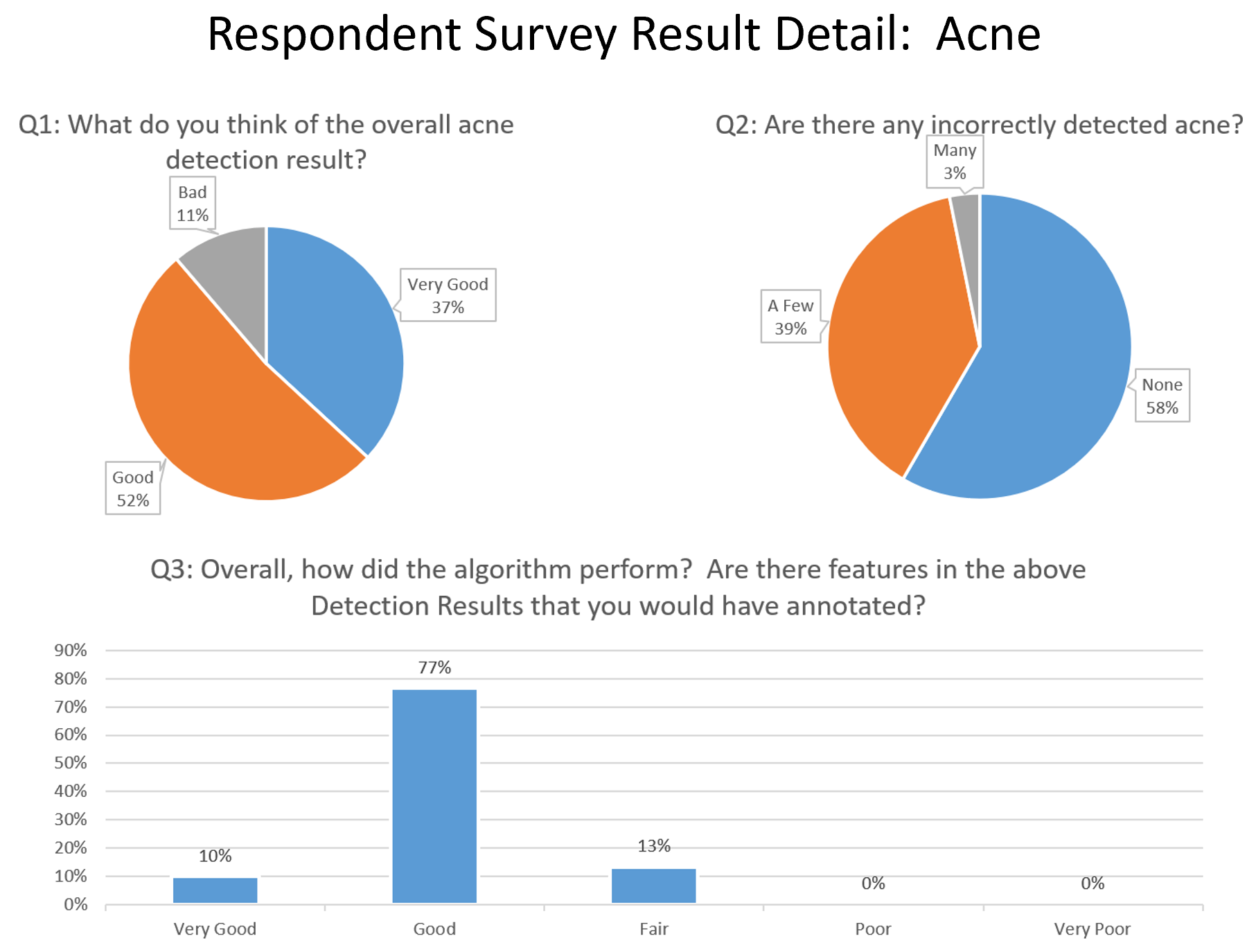}
  \caption{Details of respondent survey results for our acne detection model.}
  \label{fig:10}
\end{figure}

\noindent{\bf Acne.} As shown in Figure~\ref{fig:10}, the responses on acne detection were 37\% very good, 52\% good and 11\% bad. 
The responses on incorrectly detected acne were 58\% none, 39\% a few, and 3\% many. 
Overall, respondents felt that the acne algorithm performed well (77\% good, 10\% very good, and 13\% fair) with few images have missing detections or false positives.   
The respondents provided some additional comments regarding the acne detection results. 
Some felt that the detection for acne should be more precise. 
Some felt that acne was detected as pigmentation. 
Others felt that skin tone colour might affect the outcome of detection and image resolution can be improved to get a better result. 

\begin{figure}[!ht]
\centering
  \includegraphics[width=1\linewidth]{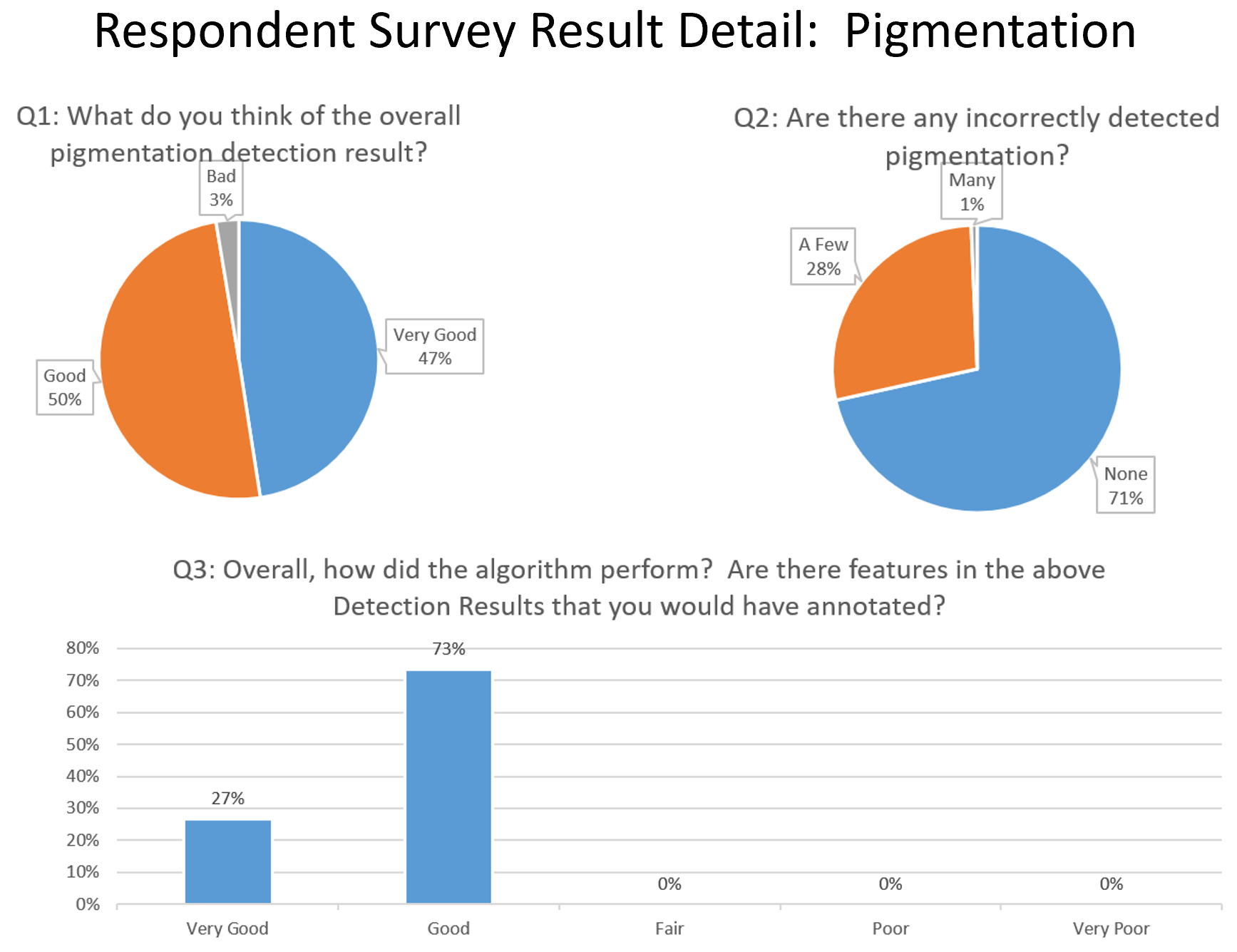}
  \caption{Details of respondent survey results for our pigmentation detection model.}
  \label{fig:11}
\end{figure}

\begin{figure}[!ht]
\centering
  \includegraphics[width=1\linewidth]{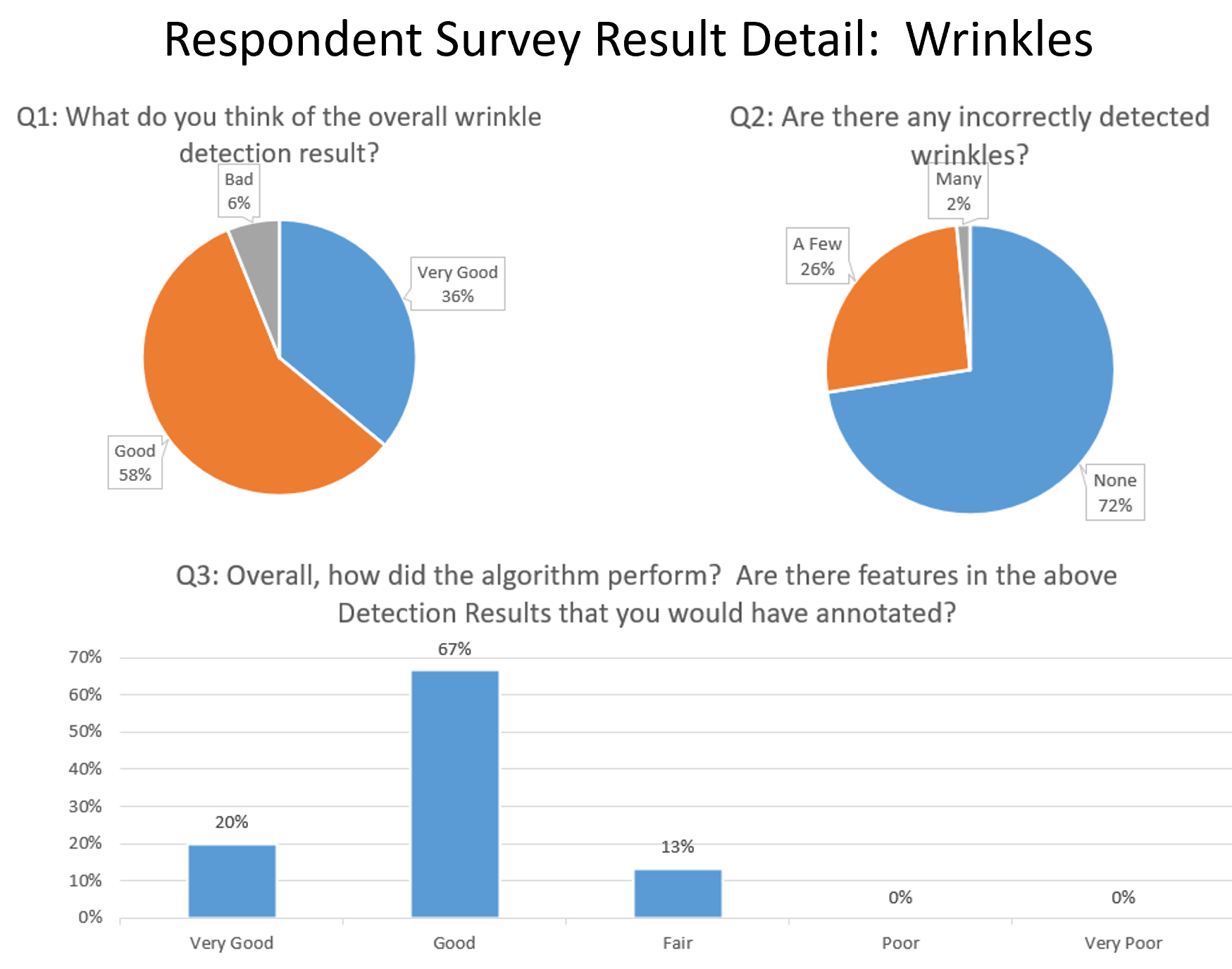}
  \caption{Details of respondent survey results for our wrinkle detection model.}
  \label{fig:12}
\end{figure}

\noindent{\bf Pigmentation.} 
As shown in Figure~\ref{fig:11}, the responses for pigmentation detection were 47\% very good, 50\% good, and 3\% bad. 
The responses on incorrectly detected pigmentation are 71\% none, 28\% a few, and 1\% many. 
Overall, respondents felt that the pigmentation algorithm performed well (73\% very good and 27\% good) with a few images have missing detections or false positives. 
In comparison between the pigmentation’s false-negative and pigmentation’s false-positive results, the perception results are better than for acne. 
The respondents provided some additional comments regarding the pigmentation detection results. 
Some felt that acne or moles was detected as pigmentation. These were not so obvious due to the image resolution and quality.

\noindent{\bf Wrinkles.}
As shown in Figure~\ref{fig:12}, the responses for wrinkle detection were 36\% very good, 58\% good, and 6\% bad. The responses on incorrectly detected wrinkles are 72\% none, 26\% a few, and 2\% many. 
Overall, respondents feel that the wrinkle algorithm performs well (67\% good, 20\% very good, and 13\% fair) with a few images have missing detection or false positives. 
The respondents added additional comments regarding the wrinkle detection results. 
Some felt that line-like features (e.g., hair, eyebrow, scars) seemed to be detected as wrinkles.

\section{Conclusion}

In this paper, we propose a deep learning based approach for the task of skin feature detection.
We focus on skin conditions of acne, pigmentation, and wrinkle. 
To train our method, we collect and annotate a large-scale training dataset. 
To achieve accurate detection, we conduct two-phase annotation to take advantage of machine learning for better annotation.
Our method produces promising results for the detection of acne, pigmentation, and wrinkle for a single selfie face in the wild.

\section{Acknowledgments}
This work was carried out at the Rapid-Rich Object Search (ROSE) Lab, Nanyang Technological University (NTU), Singapore. The research is supported by A*STAR under it’s A*STAR MBRC Strategic Positioning Fund (SPF) – A*STAR-P\&G Collaboration (Award APG2013/113). Any opinions, findings and conclusions or recommendations expressed in this material are those of the author(s) and do not reflect the views of the A*STAR.





\end{document}